# MBIC – A Media Bias Annotation Dataset Including Annotator Characteristics

Timo Spinde[1,2][0000-0003-3471-4127], Lada Rudnitckaia[1][0000-0003-0296-8376], Kanishka Sinha[3][0000-0002-1094-1707], Felix Hamborg[1][0000-0003-2444-8056], Bela Gipp[2][0000-0001-6522-3019], and Karsten Donnay[4][0000-0002-9080-6539]

[1] University of Konstanz, Germany
{firstname.lastname}@uni-konstanz.de
[2] University of Wuppertal, Germany
{last}@uni-wuppertal.de
[3] University of Passau, Germany
sinha02@ads.uni-passau.de
[4] University of Zurich, Switzerland
donnay@ipz.uzh.ch

**Abstract.** Many people consider news articles to be a reliable source of information on current events. However, due to the range of factors influencing news agencies, such coverage may not always be impartial. Media bias, or slanted news coverage, can have a substantial impact on public perception of events, and, accordingly, can potentially alter the beliefs and views of the public. The main data gap in current research on media bias detection is a robust, representative, and diverse dataset containing annotations of biased words and sentences. In particular, existing datasets do not control for the individual background of annotators, which may affect their assessment and, thus, represents critical information for contextualizing their annotations. In this poster, we present a matrix-based methodology to crowdsource such data using a self-developed annotation platform. We also present MBIC (Media Bias Including Characteristics) - the first sample of 1,700 statements representing various media bias instances. The statements were reviewed by ten annotators each and contain labels for media bias identification both on the word and sentence level. MBIC is the first available dataset about media bias reporting detailed information on annotator characteristics and their individual background. The current dataset already significantly extends existing data in this domain providing unique and more reliable insights into the perception of bias. In future, we will further extend it both with respect to the number of articles and annotators per article.

# 1   Introduction

News articles in online newspapers are considered a reliable, primary, and crucial information source that increasingly replaces traditional media like television or radio broadcasts and print media next to new information sources such as social media [2].

Research to date has widely demonstrated that news outlets can be biased [5] and that, at the same time, bias media coverage has the potential to strongly impact the public perception of the reported topics [5]. This is especially known to be relevant within the so-called "filter bubbles" or "echo chambers" [3] and can lead to poor awareness about particular issues and a narrow and one-sided point of view [9, 12]. Biased media coverage can further potentially affect the audience's political beliefs and even change voting behavior [5].

One of the key challenges of an automated media bias identification is the lack of a gold standard, large-scale dataset for labeled media bias instances. In section 2, we briefly provide an overview of existing datasets. To the best of our knowledge, all existing datasets have major drawbacks for the automated detection of bias due to their size, level of annotations, annotator characteristics, or a more limited focus on specific framing effects rather than media bias more broadly.

In comparison, our dataset covers a greater topic variety and contains both framing and epistemological bias instances. In addition, the ability to identify media bias on the word level is desired to provide concrete evidence of bias to a reader [8, 10].

# 2   Related work

Lim et al. use crowdsourcing to construct a dataset consisting of 1,235 sentences from various news articles reporting on the same event [8]. The authors argue that focusing on just one event allows capturing differences in the language used by different journalists. The dataset provides labels on the article and word level.

In another work by Lim et al., they propose another media bias dataset consisting of 966 sentences containing labels on the sentence level. The dataset covers various news about four different events: Donald Trump's statement about protesting sportsmen, Facebook data misuse, negotiations with North Korea, and a lawmaker's suicide [7].

Baumer et al. focus on the automated identification of framing in political news. Using crowdsourcing, they construct a dataset that consists of 74 news articles from various US news outlets covering diverse political issues and events [1].

Hamborg et al. constructed a dataset called *NewsWCL50* using content analysis [4]. They created a codebook describing frame properties, coding rules, and examples. The dataset consists of 50 news articles from various US news outlets and covers ten political events. The authors distinguish the target concept and phrases framing this concept. They also define some framing properties, e.g., "affection," "aggression," "other bias," and others.

Fan and White et al. created the dataset BASIL of 300 news articles covering diverse events and containing lexical and informational bias [3]. The dataset allows analysis on the token level and relative to a given target but, for lexical bias, only 448 sentences are available. The annotation was conducted by two experts.

The above datasets all have individual limitations, that can be significant drawbacks for the analysis of media bias: 1) they only include a few topics ([7], [8]), 2) they mostly focus exclusively on framing ([1], [4]), 3) annotations are target-oriented ([3], [4]), 4) annotations are not on the word level ([8]), or 5) training data are too small ([3]).

## 3    Dataset creation

### 3.1    Data collection

We created a diverse and robust dataset for media bias identification by collecting 1,700 sentences from around 1,000 articles that potentially contain media bias by word choice instances. The collection focused specifically on sentences since media bias by word choice, and labeling rarely depends on context outside the given sentence [3].

In order to cover all of the United States' political and ideological spectrum, we used articles from three left-wing media outlets: HuffPost, MSNBC, and AlterNet, three right-wing media outlets: The Federalist, Fox News, Breitbart, and two outlets from the center: USA Today and Reuters. When selecting the media outlets, we relied on media bias charts provided by Allsides[1] and Ad Fontes Media[2] and Allsides media bias ratings[3] to ascertain overall partisan leanings of each outlet [11].

Our dataset contains 14 topics that describe different events and issues that happened and were discussed in news articles from January 2019 till June 2020. We selected ten topics that are very contentious in the United States and are more likely to be described with biased language [6] (abortion, coronavirus, elections 2020, environment, gender, gun control, immigration, Donald Trump's presidency, vaccines, white nationalism). We also introduced four less contentious topics (student debt, international politics, and world news, middle class, sport) for comparison.

The collection process was as follows: We specified the keywords characterizing the selected topics, the chosen media outlets, and the Media Cloud time frame, an open-source platform for media analysis[4], to retrieve all the available links to the relevant news articles. Using the available metadata, we then manually collected the sentences with examples of media bias across the articles.

Note that we tried to include only sentences from the news section and avoid sentences from the commentary section of the selected news outlets. The ultimate goal of media bias identification systems is to recognize subtle bias arising in factual reporting – the section where, ideally, there should be no or little bias.

### 3.2    Data annotation

For annotation of the collected sentences, we engaged micro-jobbers on Amazon Mechanical Turk. Annotation quality of experts is often preferable but in this project we expressively wanted to collect a large number of annotations from non-experts. Specifically, the objective was to create data that allow insights into the perception of media bias by a broader public. We executed the study with a self-built annotation platform that allowed us to combine annotation and classic survey tasks. All instructions, questions, and the platform itself is free to use for further research on https://bit.ly/2TkJz3m.

---

[1] https://www.allsides.com/media-bias/media-bias-chart
[2] https://www.adfontesmedia.com
[3] https://www.allsides.com/media-bias/media-bias-ratings
[4] https://mediacloud.org

To the best of our knowledge, our dataset is the first in the research area to collect detailed background demographic information about the annotators, such as gender, age, education, English proficiency, but also information on political affiliation and news consumption.

In total, 784 annotators participated in the survey, all located in the United States. The vast majority (97.1%) of the annotators are native English speakers, 2.8% are near-native speakers. The annotators from diverse age groups participated in the survey; people from 20 to 40 years old prevail over other age groups. The annotators' gender is balanced between females (42.5%) and males (56.5%). The annotators have a diverse educational background; more than half have higher education. The annotators' political orientation is not well balanced: liberal annotators are in the majority as compared to conservative annotators and annotators from the center. The vast majority of the annotators read the news sometimes, more than half one or more times per day. We summarize all background information on the annotators in Fig.1.

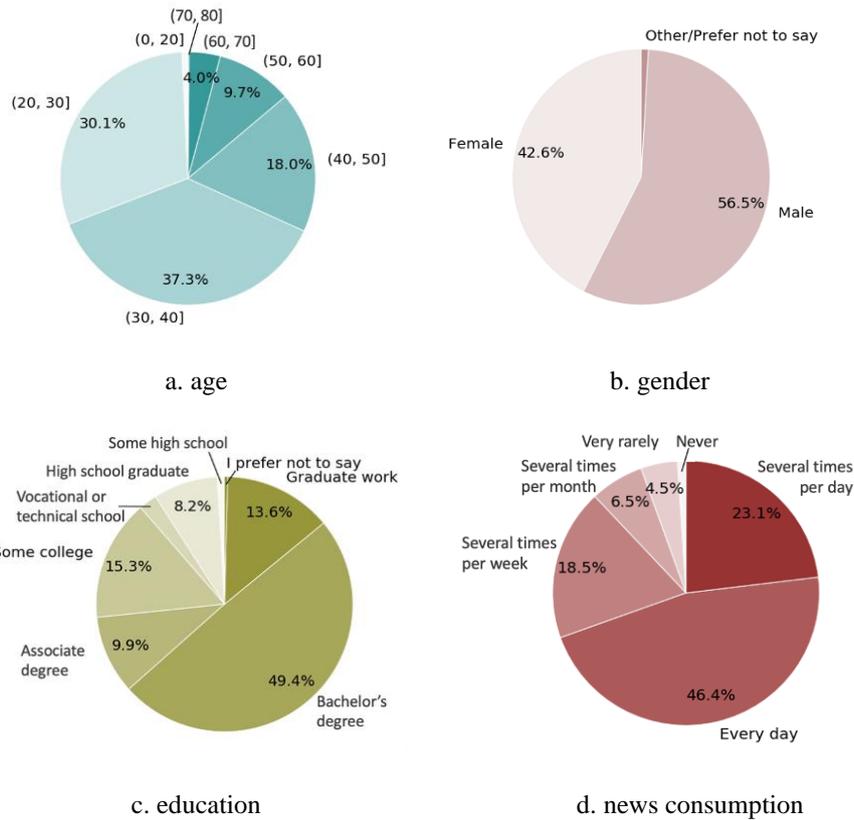

a. age

b. gender

c. education

d. news consumption

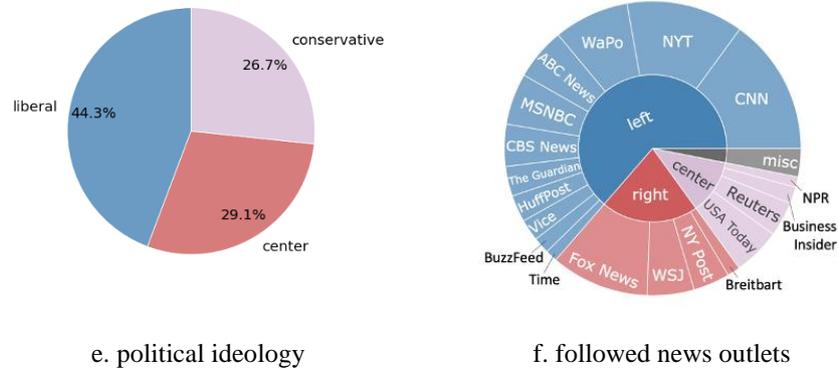

e. political ideology    f. followed news outlets

**Fig. 1.** Information about annotators: a. age, b. gender, c. education, d. news consumption, e. political ideology, f. followed news outlets

We provide detailed instructions on recognizing and annotating media bias instances and show several illustrative examples of media bias. We then ask the annotators to closely follow the instructions and leave their personal preferences aside.

After we provide the instructions, we ask one quality control question that ensures that annotators understood the instructions correctly. If an annotator answers incorrectly, she cannot proceed and is forced to reread the instructions.

First, we ask annotators to highlight words or phrases that induce bias according to the previously provided instructions. Then, we ask them to annotate the whole sentence as biased or impartial. Finally, we ask them whether the sentence is opinionated, factual, or mixed. We show the results of this classification in Fig. 2 and 3.

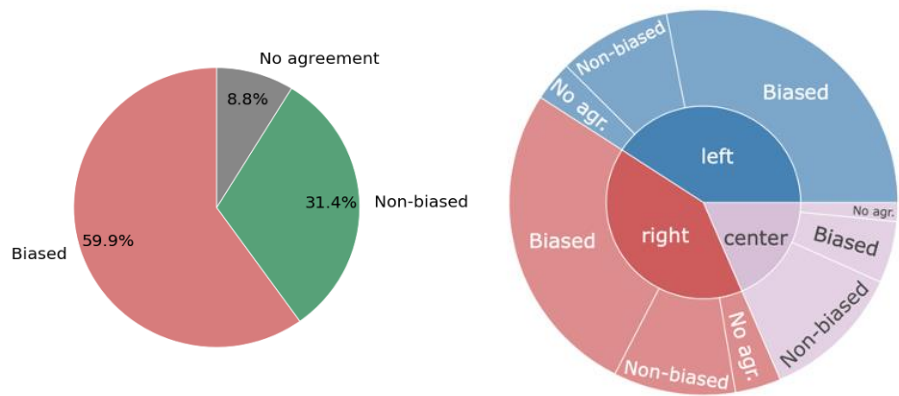

**Fig. 2.** Distribution of biased and non-biased sentences in the dataset: on the left in general, on the right per ideology of media outlets the sentences were collected from.

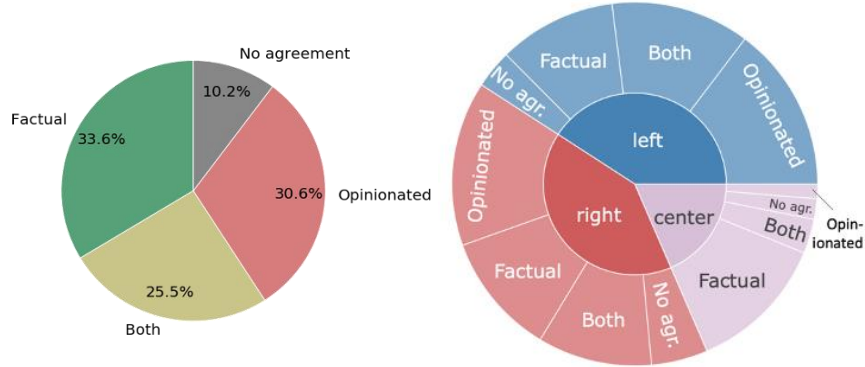

**Fig. 3.** Distribution of opinionated, factual and mixed sentences in the dataset: on the left in general, on the right per ideology of media outlets the sentences were collected from

Overall, our dataset allows performing three different tasks: bias identification on the word level, sentence level, and a classification of the sentence as being opinionated, factual or a mixture of both.

To avoid question ordering effects or interdependencies, each annotator received 20 randomly reshuffled sentences about various topics and from various outlets. The annotators did not receive any additional information about the sentences apart from the sentences themselves. We showed each sentence to ten annotators.

To motivate the workers to look for biased words more attentively and not to select all the words in the sentence, we introduced a small monetary bonus for each word that was selected by at least one other annotator and a small penalty for each selected word that was not selected by anybody else [1].

### 3.3 Final dataset characteristics

We assigned a biased or impartial label to a sentence if more than half of respondents annotated a sentence as biased or impartial, respectively. 149 sentences could not be labeled due to a lack of agreement between annotators. Fleiss Kappa for annotations whether the sentence is biased/impartial is 0.21, which can be considered as a fair agreement. It represents the general difficulty of the task: For example, Hube et al. reported $\alpha = 0.124$, and Recasens et al. reported 40.73% agreement when looking at only the most biased word in Wikipedia statements. Noteworthy, inter-rater agreement is higher between annotators who reported similar political ideology, especially within liberal annotators (see Figure 4). The annotation results confirm our data sampling strategy: biased and non-biased statements are not balanced in the dataset, biased statements prevail over non-biased statements. Besides, most media bias instances are taken from liberal and conservative news sources, whereas sources from the center were used mainly to retrieve non-biased statements. Note, that this does not imply that liberal and conservative news outlets in general experience media bias by word choice and labeling and provide opinionated news more often than news outlets from the center. We observe these differences due to our data collection scheme.

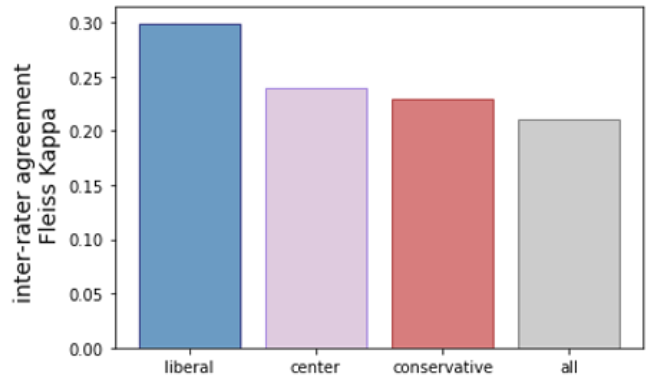

**Fig. 4.** Differences in inter-coder agreement between annotators with different political ideology

We assigned an opinionated, factual, or mixed label to a sentence if most respondents annotated a sentence as opinionated, factual, or mixed, respectively. 174 sentences could not be labeled due to a lack of agreement between annotators. According to our crowdsourced annotations, the dataset contains an almost equal number of factual, opinionated, and mixed statements.

The annotation scheme for biased words allowed respondents to highlight not only the words but also short phrases. A word was considered biased on the word level if at least four respondents highlighted it as biased. On average, a sentence that contains biased words contains two biased words. Out of 31,794 words for training, only 3,018 are biased, which forms only 9.5% of our current data.

## 4  Conclusion and further work

We present MBIC, a diverse dataset containing annotations of biased words and sentences, crowdsourced by non-experts while asking and reporting their background in detail. We argue that the research community lacks large labeled datasets for use in media bias detection methods. We also believe that the data could be interesting for other research areas, especially since they measure the perception of bias by a broad and diverse public audience. The articles in our dataset include a variety of topics, from controversial to non-controversial, and recent as well as general topics. We publish the full data set at https://zenodo.org/record/4474336#.YBHO6xYxmK8.

In future work, we will further extend the data with annotations by both expert and non-expert annotators and report on the differences, respectively the overlap, of both groups. We will also evaluate whether requiring more than ten annotators per sentence leads to a significant performance increase and which amount of agreement could be seen as high quality for the area. Lastly, we will perform an additional study to control for the influence of the exact wording of our questions on the perceptions of annotators and develop guidelines for conducting such an annotation task.

**How to cite this paper:**
T. Spinde, L. Rudnitckaia, K. Sinha, F. Hamborg, B. Gipp, K. Donnay "MBIC – A Media Bias Annotation Dataset Including Annotator Characteristics". In: Proceedings of the iConference 2021.

**BibTex:**
```
@InProceedings{Spinde2021MBIC,
  title = {MBIC – A Media Bias Annotation Dataset Including Annotator Characteristics},
  booktitle = {Proceedings of the iConference 2021},
  author = {Spinde, Timo and Rudnitckaia, Lada and Sinha, Kanishka, and Hamborg, Felix and and Gipp, Bela and Donnay, Karsten},
  year = {2021},
  location  = {Beijing, China (Virtual Event)},
  month = {March},
  topic = {newsanalysis},
}
```